%% file: emnlp2021.tex
\pdfoutput=1

\documentclass[11pt]{article}

\usepackage{emnlp2021}

\usepackage{times}
\usepackage{latexsym}
\usepackage{epsfig}
\usepackage{graphicx}
\usepackage{xcolor}
\usepackage{amsmath}
\usepackage{amssymb}
\usepackage{multicol}
\usepackage{multirow}
\usepackage[caption=false]{subfig}
\usepackage[T1]{fontenc}
\usepackage{comment}
\usepackage[utf8]{inputenc}

\usepackage{microtype}
\usepackage[hyphenbreaks]{breakurl}

\title{CIDEr-R: Robust Consensus-based Image Description Evaluation}

\author{
  Gabriel Oliveira dos Santos \and Esther Luna Colombini \and Sandra~Avila \\
  Institute of Computing, University of Campinas (Unicamp), Brazil\\
  \texttt{{g197460@dac.unicamp.br, \{esther,sandra\}@ic.unicamp.br}}
 }

\begin{document}
\maketitle

\begin{abstract}
\input{00-abstract}

\end{abstract}
\input{01-introduction}
\input{02-related_work}
\input{03-dataset_summary}

\input{04-model}
\input{05-metrics}
\input{06-experiments}
\input{07-conclusion}

\section*{Acknowledgements}
G. O. dos Santos is funded by the São Paulo Research Foundation (FAPESP) (2019/24041-4)\footnote{The opinions expressed in this work do not necessarily reflect those of the funding agencies.}. E. L. Colombini and S. Avila are partially funded by H.IAAC (Artificial Intelligence and Cognitive Architectures Hub). S. Avila is also partially funded by~FAPESP (2013/08293-7), a CNPq PQ-2 grant (315231/2020-3), and Google LARA~2020. 

\bibliography{bibliography}
\bibliographystyle{acl_natbib}




\end{document}

%% file: 00-abstract.tex
This paper shows that CIDEr-D, a traditional evaluation metric for image description, does not work properly on datasets where the number of words in the sentence is significantly greater than those in the MS COCO Captions dataset. We also show that CIDEr-D has performance hampered by the lack of multiple reference sentences and high variance of sentence length. To bypass this problem, we introduce CIDEr-R, which improves \text{CIDEr-D}, making it more flexible in dealing with datasets with high sentence length variance. We demonstrate that CIDEr-R is more accurate and closer to human judgment than CIDEr-D; CIDEr-R is more robust regarding the number of available references. 
Our results reveal that using Self-Critical Sequence Training to optimize CIDEr-R generates descriptive captions. In contrast, when CIDEr-D is optimized, the generated captions' length tends to be similar to the reference length. However, the models also repeat several times the same word to increase the sentence length.

%% file: 01-introduction.tex
\section{Introduction}
\label{sec:introduction}
 
Automatically describing image content using natural sentences is an important task to help include people with visual impairments on the Internet, making it more inclusive and democratic. This task is known as \emph{image captioning}.  It is still a big challenge that requires understanding the objects, their attributes, and the actions they are involved in. Thus, linguistic models are also needed to verbalize the semantic relations in addition to visual interpretation methods.

\begin{figure}[t]
\centering
    \subfloat[\textbf{Reference}: ``Fotografia aérea sobre o pedágio da Terceira Ponte. A foto contém alguns prédios, um pedaço da Terceira Ponte e o fluxo de carros.''\\
    \textbf{CIDEr-D}: ``foto \textcolor{blue}{\textit{aérea}} \textcolor{blue}{\textit{aérea}} \textcolor{blue}{\textit{aérea}} da cidade de Florianópolis mostrando \textcolor{blue}{\textit{casas}} \textcolor{blue}{\textit{casas}}, mostrando \textcolor{blue}{\textit{casas}} \textcolor{blue}{\textit{casas}}. Ao fundo, algumas \textcolor{blue}{\textit{casas}} e \textcolor{blue}{\textit{casas}}.''\\
    \textbf{CIDEr-R}: ``Foto aérea de uma cidade de Brasília. Ao fundo, a céu azul.'']{\includegraphics[trim={0 1.5cm 0 0},clip,  width=0.47\textwidth]{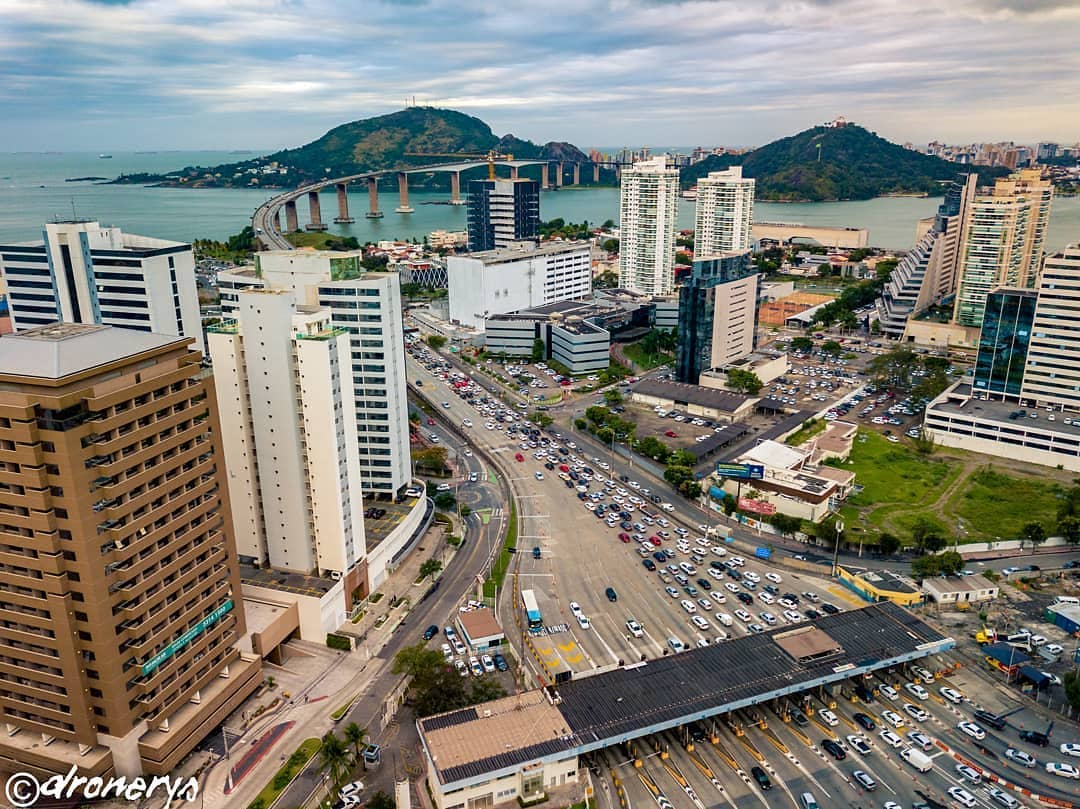}}
    \vspace{-0.1cm}
    \caption{We show the reference caption and the description generated by models trained on \#PraCegoVer to optimize CIDEr-D and CIDEr-R.}
    \label{fig:generated_descriptions_city1}
\end{figure} 

In the last few years, there has been significant progress in image captioning thanks to the availability of a large amount of annotated data in relevant datasets \cite{agrawal2019nocaps,chen2015microsoft,gurari2020captioning,Flickr8K2013, FlicKr30K2017}. These datasets have in common multiple reference captions, which comprehend short sentences with a low length variance. Also, most of the datasets contain only English captions, whereas datasets with captions described in other languages are scarce. Thus, we proposed the \#PraCegoVer dataset \cite{santos2021pracegover}
for the image captioning task with descriptions\footnote{We use caption and description interchangeably.} in Portuguese. Our dataset consists of images and captions collected from posts on Instagram tagging the hashtag \textit{\#PraCegoVer}. \textit{PraCegoVer} is a movement \cite{criadoraPracegover} that aims to spread the accessibility culture on social media. It has inspired many local laws in Brazil that establish that all posts made by public agencies must tag \emph{\#PraCegoVer} and contain a short description of~the image. Also, many business and personal profiles have joined this initiative. Hence, the amount of data produced by those users has great growth potential.

In contrast to the other datasets in this literature, in \#PraCegoVer, each image has only one reference description, whose length is about four times longer than the average length, in terms of the number of words, in datasets such as MS COCO \cite{chen2015microsoft}. Also, the variance of length in our dataset is around ten times higher than in the literature. Thus, \#PraCegoVer is more challenging than others. We demonstrated that training models using Self-Critical Sequence Training \cite{rennie2017self} (SCST) to optimize CIDEr-D is not a good approach for datasets with the same characteristics as ours, in particular with single reference and high variance of length. \text{CIDEr-D}~\cite{CIDEr2015} has a severe penalty concerning the length that punishes at the same intensity any difference in the number of words between the generated and the reference captions. Hence, it forces the model to generate descriptions similar in length to the ground truth even though it has to repeat words to reach the reference length. In the end, the model learns to repeat words impairing the lexical cohesion of the generated description.

We tackle the problem of penalizing at the same intensity any difference in length by modeling the penalty proportionally to the reference length. Also, we introduced a penalty regarding lexical cohesion so that the more a word is repeated, relative to the ground truth, the higher is the penalty. These two penalties combined aim to give the models the opportunity of generating more succinct captions when the ground truth comprehends longer descriptions. We name this new metric CIDEr-R, a robust alternative to CIDEr-D for datasets with high length variance. Also, we demonstrated that CIDEr-R is preferred to CIDEr-D when only one reference is available. Finally, our results reveal~that CIDEr-R is more suitable for the context of datasets created from data collected from social media.

Our key contributions are fourfold: 
    (1) we investigated the impact of evaluation metrics in the scenario of datasets with high variance of length (\textit{e.g.}, \#PraCegoVer; 
    
    (2) we showed that optimizing \text{CIDEr-D} on datasets with long references and high variance of length can make the model learn to repeat words to increase the sentence length (Figure~\ref{fig:generated_descriptions_city1});
    
    (3) we proposed CIDEr-R that better measures the lexical cohesion and has a flexible penalty to length variance;
    
    (4) we demonstrated that optimizing CIDEr-R in context of long sentences and high variance of length produce more descriptive captions.

%% file: 02-related_work.tex
\section{Related Work}
\label{sec:related_works}

The image captioning task has been accelerated thanks to the availability of a large amount of annotated data in relevant datasets, for instance, Flickr8K~\cite{Flickr8K2013}, Flickr30K~\cite{FlicKr30K2017}, and MS COCO~\cite{MSCoco2014}. These data made it possible to train complex models such as Deep Neural Networks~\cite{lecun2015deep}. In the last few years, a broad collection of architectures has been proposed in image captioning \cite{Kulkarni11babytalk, vinyals2015show, johnson2016densecap, pedersoli2017areas, anderson2018bottom, Lu_2018_CVPR, AoANet_2019, wang2020visual, cornia2020meshed, pan2020x, li2020oscar}.  Notable achievements have been made with \textit{Self-Critical Sequence Training}~\cite{rennie2017self}, which is a formulation of reinforcement learning that enables the use of non-differentiable functions as optimization objectives, such as caption metrics.

Evaluating the generated captions is a challenging task itself because it is based on the evaluator's subjectivity. Many metrics were proposed to measure the generated text's quality to encourage progress in this field despite the subjectivity intrinsic to this process. Typically, these metrics compare a candidate sentence against a set of reference sentences (ground truth). Most of the metrics are rule-based that relies on n-gram to assign a score to the candidate sentence, \textit{e.g.} BLEU \cite{BLEU2002}, ROUGE \cite{ROUGE2004} and METEOR \cite{METEOR2005}. However, n-gram based metrics cannot capture the semantic relations and negatively correlate with human judgment. Thus, Vedantam \textit{et al.} proposed CIDEr \cite{CIDEr2015}, a semantically sensitive score, which computes the term frequency-inverse document frequency (TF-IDF) weights to n-grams in the candidate and reference sentences and then compare them using cosine similarity. SPICE \cite{anderson2016spice} uses the similarity of scene graphs parsed from the reference and candidate captions to evaluate the generated descriptions. Although SPICE obtains a significantly higher correlation with human judgments compared to n-gram based metrics, it does not check whether the grammar is correct, and it ignores syntactic quality, as pointed out in \cite{liu2017improved}. Also, SPICE uses Stanford Scene Graph Parser \cite{schuster2015generating} which is only available for a limited number of languages. Recently, word embeddings has also been employed for evaluation in  BERTScore \cite{zhang2019bertscore}, MoverScore \cite{zhao-etal-2019-moverscore} and BLEURT \cite{sellam-etal-2020-bleurt}.

%% file: 03-dataset_summary.tex
\section{\#PraCegoVer Dataset}
\label{sec:dataset_summary}

Recently, a social movement called PraCegoVer \cite{criadoraPracegover} arose on the Internet, standing for people with visual impairments besides having an educational purpose. The initiative aims to call attention to the accessibility question by stimulating users to post images accompanied by a short description of their content. This project has inspired many local laws that establish that public agencies' posts on social media must contain a description of the image.

Inspired by this movement, we developed a crawler to collect public posts from Instagram associated with the tag used by the followers. We leverage these data, which comprise images and text captions, to create \#PraCegoVer \cite{santos2021pracegover} dataset, the first large dataset for image captioning in Portuguese. Similar to VizWiz-Captions \cite{gurari2020captioning}, our dataset's captions are addressed to visually impaired people. We are continuously collecting more data, and, thus, our dataset increases over time. Currently, it comprehends 551,000 instances with images labeled with a single caption.

\subsection{\#PraCegoVer \textit{vs.} MS COCO Captions}

\begin{figure}[t]
    \includegraphics[width=0.49\textwidth, trim=0.2cm 0 0 0, clip]{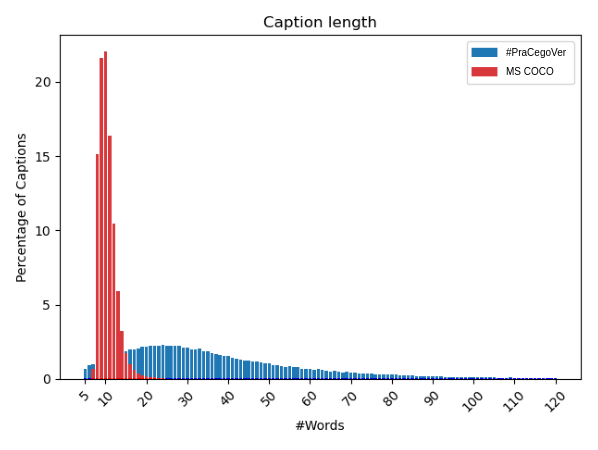}\vspace{-0.1cm}
    \caption{Histogram with the distribution of captions by length in terms of number of words for \#PraCegoVer (blue) and MS COCO Captions (red) datasets.}
    \label{fig:histogram_description_ditribuition}
    \centering
\end{figure}

To illustrate the differences between standard datasets and datasets created from data in the wild, we plot in Figure \ref{fig:histogram_description_ditribuition} a histogram with the distribution of captions by length, \textit{i.e.} number of words, for \#PraCegoVer and MS COCO Captions. We can note that in MS COCO Captions the caption lengths are concentrated around ten words, while length distribution in \#PraCegoVer is flatter than MS COCO Captions. More precisely, the reference descriptions in MS COCO Captions have 10.5 words on average with a standard deviation equal to 2.2, whereas in \#PraCegoVer the mean and standard deviation the lengths are 37.8 and 26.8, respectively. These characteristics make our dataset more challenging than MS COCO Captions because the most used evaluation metric in image captioning literature, CIDEr-D \cite{CIDEr2015}, is fine-tuned on standard datasets, then it can not evaluate the generated captions appropriately.

%% file: 04-model.tex
\section{Model Architecture}
\label{sec:model}

In our experiments, we employ AoANet \cite{AoANet_2019} since it is one of the state-of-art models for image captioning. This model is based on the Attention on Attention (AoA) module, which determines the importance of attention results given queries. AoA module is an extension of  Multi-Head Attention~\cite{attention_is_all_you_need_2017} that introduces an ``Attention Gate'' and an ``Information Vector'' to address the issue of the attention mechanism returning irrelevant results to the query. The model is an Encoder-Decoder architecture where the encoder module comprises the Feature Extractor and a Refining Module to refine feature vectors' representation. In contrast, the decoder network comprehends LSTM, Multi-Head Attention, Attention on Attention, and word prediction layers. The encoder takes the image features as input and creates an embedding representation. The decoder, in turn, uses the embedding and the previous state to model the context vector that represents the newly acquired information. The context vector is used to predict the next word.

We train our models using the method Self-Critical Sequence Training \cite{rennie2017self}, which consists of two phases. The first phase optimizes the cross-entropy loss, and then an evaluation metric is optimized through a Reinforcement Learning approach. Our experiments trained the models during two epochs, minimizing the cross-entropy loss and 15 epochs maximizing the considered evaluation metric.

%% file: 05-metrics.tex
\section{CIDEr-R: Our Proposal}
\label{sec:cider_modified}

CIDEr-D~\cite{CIDEr2015}  is based on vectors $\mathbf{g^n}(s)$ formed by \text{TF-IDF} scores computed from $n$-grams of sentence $s$. CIDEr-D can be splitted into terms $\text{CIDEr-D}_n$, which are scores calculated from $n$-grams of length $n$. Let $S_i = \{s_{i1}, ... ,s_{im}\}$ be the set of reference descriptions of an image $i$, and let $c_i$ be the generated description for $i$. Then, \text{CIDEr-D}$_n$  is formally defined as follows:
  
 \begin{equation}
 \small
  sim(c_i, s_{ij}) = \frac{ min(\mathbf{g^n}(c_i), \mathbf{g^n}(s_{ij})) \cdot \mathbf{g^n}(s_{ij})}{||\mathbf{g^n}(c_i)||\cdot||\mathbf{g^n}(s_{ij})||}, 
 \end{equation}
 
 \begin{equation}
 \small
  Penalty = exp\{-\frac{(l(c_i)-l(s_{ij}))^2}{2\sigma^2}\},
 \end{equation}
 
 \begin{equation}
 \label{eq.cider_d}
 \small
     \text{CIDEr-D}_n(c_i, S_i) = \frac{10}{m} \sum_{s_{ij}\in S_i} sim(c_i, s_{ij}) \cdot Penalty,
 \end{equation}
 
\noindent where $l(c_i)$ and $l(s_{ij})$ denote the length of candidate and reference sentences, respectively, and it is used $\sigma=6$.
 
 Finally, the CIDEr-D score is defined as:
 \begin{equation}
 \small
     \text{CIDEr-D}(c_i, S_i) = \frac{1}{N} \sum_{n=1}^{N}\text{CIDEr-D}_n(c_i, S_i),
 \end{equation}
 
\noindent where $N$ is the longest considered $n$-gram, in~\cite{CIDEr2015} is told that empirically is assigned $N=4$.\vspace{0.1cm}

Although Vedantam \textit{et al.}~\cite{CIDEr2015} have introduced the Gaussian penalty $exp\{-\frac{(l(c_i)-l(s_{ij}))^2}{2\sigma^2}\}$ to tackle the problem of higher scores when words with higher confidence are repeated over long sentences, this factor only works for datasets with slight variances in the sentence lengths, and where the sentences have few words. For instance, this is the case of MS COCO, but it is not the case of the \#PraCegoVer. We have demonstrated in Section~\ref{sec:experiments} that models based on one of the state-of-art architectures trained on MS COCO and optimizing CIDEr-D output descriptions with high quality. However, when we trained these models with the same hyperparameters over the \#PraCegoVer, the result was poor, with output sentences including many word repetitions. Such behavior occurs because the Gaussian penalty punishes at the same intensity any difference in length $|l(c_i)-l(s_{ij})|$. Hence, the model is forced to generate captions similar in length to the reference, learning to repeat words not to be penalized by the cosine similarity between the TF-IDF vectors.

In light of this problem, we propose CIDEr-D modified, which we called CIDEr-R, where we replace the Gaussian penalty with two other penalties: length and repetition. We relax the Gaussian Penalty to be more flexible in the context of varied sentence length. Therefore, we developed the \textbf{length penalty} that changes according to the reference length so that the longer the reference sentence, the more permitted is the difference $|l(c_i)-l(s_{ij})|$. Also, to avoid the model to predict repeated words, we introduce the \textbf{repetition penalty} that deals specifically with this problem. This penalty considers the number of occurrences of a word in the candidate and reference sentences, assigning an intensified penalty score as that word frequency in the candidate sentence differs from the frequency in the reference.

Formally, given the generated description $c_i$ and the reference $s_{ij}$, the \textbf{length penalty} $Pen_{L}(c_i, s_{ij})$ is defined as follows:
\begin{equation}
\label{eq.pen_length}
\small
    Pen_{L}(c_i, s_{ij}) =  exp\{-\frac{(l(c_i)-l(s_{ij}))^2}{l(s_{ij})^2}\}.
\end{equation} 

\textbf{Repetition penalty} $Pen_{R}(c_i, s_{ij})$ is formally defined as:
\begin{equation}
\label{eq.pen_repetition}
\small
    Pen_{R}(c_i, s_{ij}) =   \prod_{w \in c'_i} f(w, c_i, s_{ij})^{1/l(c_i)},
\end{equation} 
\begin{equation*}
\scriptsize
    f(w, c_i, s_{ij}) = \begin{cases}\frac{1}{1+|freq(w, c_i) - freq(w, s_{ij})|} \mbox{, if } w \in c'_i \cap s'_{ij}  \\
                      \frac{1}{freq(w, c_i)} \hspace{2cm} \mbox{, if } w \in c'_i \setminus s'_{ij} 
                      \end{cases}
\end{equation*}  

\noindent where $freq(w, s)$ is the number of occurrences of the word $w$ in a sentence $s$, and the notation $c'$ and $s'$ represents the set of words that appear in $c$ and $s$, respectively.

Note that the length penalty (Equation~\ref{eq.pen_length}) is similar to the Gaussian penalty in Equation~\ref{eq.cider_d}, except for the denominator, which we replace by the term $l(s_{ij})^2$. This change makes the metric more flexible concerning the difference in lengths generated and reference sentences.

Finally, we define CIDEr-R as follows:
 \begin{equation}
 \label{eq.cider_d_modified}
 \scriptsize
     \text{CIDEr-R}_n(c_i, S_i) =
     \frac{10}{m} \sum_{s_{ij}\in S_i} sim(c_i, s_{ij}) \cdot Pen_{R}^{k_r} \cdot Pen_{L}^{1-k_r},
 \end{equation}
 \begin{equation*}
 \small
     k_r \in [0, 1].
 \end{equation*}
 
Our formulation of CIDEr-R has replaced the Gaussian penalty with the weighted geometric average of length and repetition penalties, where $k_r$ and $1-k_r$ is the weight assigned to repetition and length penalty, respectively.

We have executed a random search~\cite{bergstra2012random} to tune the weights assigned to each penalty, and we find the best configuration is $k_r = 0.8$. This configuration forces the model to avoid repetition. On the other hand, it generates shorter sentences than when $k_r = 0.0$, which is expected since we are decreasing the length penalty's weight.

%% file: 06-experiments.tex
\section{Experiments}
\label{sec:experiments}

\subsection{Caption-Level Human Correlation}
\label{sub:human_level_correlation}
In this experiment, we evaluate how much the automated metrics match human judgment. We use the PASCAL-50S proposed by Vedantam \textit{et al.} \cite{CIDEr2015}. This dataset consists of triplets of sentences A, B, and C, where A is a reference caption and B and C are candidates. Each triplet has an annotation with the majority vote for the question, ``Which of the two sentences, B or C, is more similar to A?''. The dataset comprehends four different combinations of sentences B and C: human-correct (HC), human-incorrect (HI), human-model (HM), and model-model (MM). 

\begin{figure}[t]
    \includegraphics[width=0.485\textwidth, trim=0.2cm 0 0 0, clip]{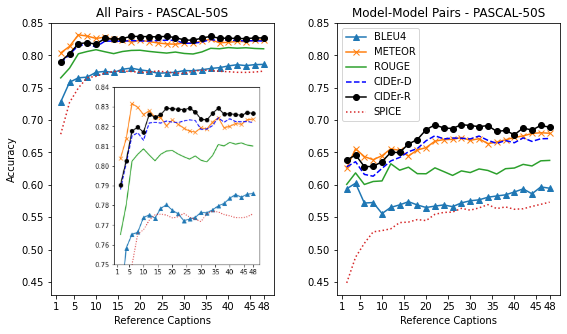}
    \caption{Classification accuracy of metrics at matching human evaluation with 1-48 reference captions.}
    \label{fig:metrics_experiment_results}
    \centering
\end{figure}

To evaluate the metrics, we compute the score for each triplet, using different metrics, for B and C having a set of references, and then we calculate the metric's accuracy. The accuracy is defined as the number of times the metric assigns the highest score to the candidate pair most commonly preferred by human evaluators divided by the total number of pairs (B, C). To assess the impact of the number of available references on metrics performance, we compute the accuracy for each metric, varying the number of available references from 1 to 48. We show the results in Figure \ref{fig:metrics_experiment_results}\footnote{We used the implementation of the evaluation metrics available on \url{https://github.com/ruotianluo/coco-caption}}.

We can see that CIDEr-R has a similar performance to CIDEr-D, considering the four types of pairs. However, CIDEr-R outperforms the other metrics at distinguishing between two machine-generated captions (Model-Model Pairs). Especially when only one reference caption is available, which is the most common scenario of large datasets such as ours, CIDEr-R outperforms CIDEr-D by 1.2\% of accuracy. Model-Model Pairs is a most interesting subtask from a practical point of view because distinguishing better-performing algorithms is the primary motivation of the automated metrics.

\subsection{Ablation Study}

To investigate the impact of mean and variance of the sentence's length in the datasets and influence of availability of multiple references, we selected subsets from \#PraCegoVer and MS COCO datasets. We constructed MS COCO -- Subset (Single Reference) to increase the sentence length variance. Thus, we selected the longest reference for each image, and then we removed the examples whose sentence length is between the first and third quartiles. Regarding the MS COCO -- Subset (Multi-Reference), we kept the same examples as in MS COCO -- Subset (Single Reference) but considered the 5 references per image. In addition, we randomly selected a subset of \#PraCegoVer with about 63 thousand examples, which we called \textit{\#PraCegoVer-63K}. Also, we constructed the subsets \textit{\#PraCegoVer N Words} aiming to reduce the variance of the sentence length, so we selected just the examples whose reference has at most N-words. Finally, we used the original datasets MS COCO and Flickr30k. We presented the statistics about the datasets used in our experiments in Table \ref{tab:dataset_stats}.

Self-Critical Sequence Training has been largely used to train the Image Captioning models. It can optimize non-differentiable functions. Therefore it is used to optimize CIDEr-D. To compare the performance of models optimizing CIDEr-D and CIDEr-R, we train models using AoANet architecture, which is still one of the state-of-the-art architectures, on the different datasets mentioned above. For each setting, we execute the training 3 times, and we present the experimental results in Table \ref{tab:experimental_results}. Note that Stanford Scene Graph Parser is not available for Portuguese, then we could not compute the SPICE score for subsets of\text{ \#PraCegoVer}.

\begin{table*}[h]
\centering
\caption{Statistics of each dataset used in our experiment. ``Avg. Sentence Length'' stands for the average sentence length, and ``Std. Sentence Length'' stands for the standard deviation of the sentence length.}\vspace{-0.1cm}
\label{tab:dataset_stats}
\resizebox{0.7\textwidth}{!}{%
\begin{tabular}{|crrrrrrr|}
\hline
\textbf{Dataset} &
  \textbf{\begin{tabular}[r]{@{}r@{}}Dataset \\ Size\end{tabular}} &
  \textbf{\begin{tabular}[r]{@{}r@{}}Train\\ Size\end{tabular}} &
  \textbf{\begin{tabular}[r]{@{}r@{}}Validation\\ Size\end{tabular}} &
  \textbf{\begin{tabular}[r]{@{}r@{}}Test\\ Size\end{tabular}} &
  \textbf{\begin{tabular}[r]{@{}r@{}}Vocabulary \\ Size\end{tabular}} &
  \textbf{\begin{tabular}[r]{@{}r@{}}Avg. Sentence\\  Length\end{tabular}} &
  \textbf{\begin{tabular}[r]{@{}r@{}}Std. Sentence \\ Length\end{tabular}} \\ \hline
MS COCO                                                                        & 123287 & 113287 & 5000  & 5000  & 13508 & 10.5 & 2.2  \\
\begin{tabular}[c]{@{}c@{}}MS COCO -- Subset \\ (Multi-Reference)\end{tabular}  & 77719  & 71428  & 3145  & 3146  & 7985  & 10.6 & 2.8  \\
\begin{tabular}[c]{@{}c@{}}MS COCO -- Subset \\ (Single Reference)\end{tabular} & 77719  & 71428  & 3145  & 3146  & 4365  & 13.7 & 3.9  \\ \hline
Flickr30k                                                                      & 31014  & 29000  & 1014  & 1000  & 18459 & 12.3 & 5.2  \\ \hline
\#PraCegoVer -- 63K                                                     & 62935  & 37881  & 12442 & 12612 & 55029 & 37.8 & 26.8 \\
\#PraCegoVer 10 Words  & 10616  & 8531   & 1007  & 1078  & 5220  & 7.9  & 1.7  \\
\#PraCegoVer 20 Words  & 42292  & 33892  & 3521  & 4879  & 18829 & 13.8 & 4.3  \\ \hline
\end{tabular}%
}
\end{table*}

\begin{table*}[h]
\centering
\caption{Experimental results obtained with AoANet models trained on different datasets, using Self-Critical Sequence Training to optimize CIDEr-D and CIDEr-R. 
The results are presented as ``mean score $\pm$ standard deviation of the scores''.}\vspace{-0.1cm}
\label{tab:experimental_results}
\resizebox{0.7\textwidth}{!}{%
\begin{tabular}{|ccrrrrrr|}
\hline
\textbf{Dataset} &
  \textbf{Optimizing} &
  \textbf{CIDEr-R} &
  \textbf{CIDEr-D} &
  \textbf{SPICE} &
  \textbf{ROUGE-L} &
  \textbf{METEOR} &
  \textbf{BLEU-4} \\ \hline
\multirow{2}{*}{MS COCO} &
  CIDEr-D &
  121.6 \scriptsize{$\pm$ 0.3} &
  120.5 \scriptsize{$\pm$ 0.3} &
  21.1 \scriptsize{$\pm$ 0.1} &
  57.5 \scriptsize{$\pm$ 0.2} &
  27.7 \scriptsize{$\pm$ 0.0} &
  36.5 \scriptsize{$\pm$ 0.1} \\
 &
  CIDEr-R &
  124.2 \scriptsize{$\pm$ 0.7} &
  116.2 \scriptsize{$\pm$ 1.0} &
  20.3 \scriptsize{$\pm$ 0.0} &
  57.2 \scriptsize{$\pm$ 0.2} &
  26.8 \scriptsize{$\pm$ 0.2} &
  35.9 \scriptsize{$\pm$ 0.3} \\ \hline
\multirow{2}{*}{\begin{tabular}[c]{@{}c@{}}MS COCO -- Subset \\ (Multi-Reference)\end{tabular}} &
  CIDEr-D &
  117.6 \scriptsize{$\pm$ 0.2} &
  115.3 \scriptsize{$\pm$ 0.4} &
  20.3 \scriptsize{$\pm$ 0.1} &
  56.7 \scriptsize{$\pm$ 0.1} &
  27.0 \scriptsize{$\pm$ 0.1} &
  34.9 \scriptsize{$\pm$ 0.2} \\
 &
  CIDEr-R &
  119.5 \scriptsize{$\pm$ 0.5} &
  108.5 \scriptsize{$\pm$ 1.0} &
  19.2 \scriptsize{$\pm$ 0.2} &
  56.0 \scriptsize{$\pm$ 0.2} &
  25.6 \scriptsize{$\pm$ 0.2} &
  33.3 \scriptsize{$\pm$ 0.4} \\ \hline
\multirow{2}{*}{\begin{tabular}[c]{@{}c@{}}MS COCO -- Subset \\ (Single Reference)\end{tabular}} &
  CIDEr-D &
  101.0 \scriptsize{$\pm$ 0.4} &
  97.9 \scriptsize{$\pm$ 1.0} &
  22.7 \scriptsize{$\pm$ 0.2} &
  36.8 \scriptsize{$\pm$ 0.1} &
  16.7 \scriptsize{$\pm$ 0.1} &
  11.0 \scriptsize{$\pm$ 0.1} \\
 &
  CIDEr-R &
  110.5 \scriptsize{$\pm$ 1.4} &
  93.5 \scriptsize{$\pm$ 1.8} &
  23.5 \scriptsize{$\pm$ 0.2} &
  36.2 \scriptsize{$\pm$ 0.1} &
  15.6 \scriptsize{$\pm$ 0.1} &
  9.5 \scriptsize{$\pm$ 0.3} \\ \hline
\multirow{2}{*}{Flickr30k} &
  CIDEr-D &
  51.5 \scriptsize{$\pm$ 0.3} &
  45.3 \scriptsize{$\pm$ 0.2} &
  12.2 \scriptsize{$\pm$ 0.3} &
  44.0 \scriptsize{$\pm$ 0.2} &
  18.3 \scriptsize{$\pm$ 0.1} &
  21.7 \scriptsize{$\pm$ 0.3} \\
 &
  CIDEr-R &
  52.9 \scriptsize{$\pm$ 0.4} &
  44.2 \scriptsize{$\pm$ 0.9} &
  11.9 \scriptsize{$\pm$ 0.4} &
  44.0 \scriptsize{$\pm$ 0.7} &
  17.8 \scriptsize{$\pm$ 0.3} &
  21.1 \scriptsize{$\pm$ 0.4} \\ \hline
\multirow{2}{*}{\#PraCegoVer -- 63K} &
  CIDEr-D &
  10.8 \scriptsize{$\pm$ 1.8} &
  4.7 \scriptsize{$\pm$ 0.7} &
  - &
  14.5 \scriptsize{$\pm$ 0.4} &
  7.1 \scriptsize{$\pm$ 0.1} &
  1.6 \scriptsize{$\pm$ 0.2} \\
 &
  CIDEr-R &
  12.6 \scriptsize{$\pm$ 0.9} &
  3.3 \scriptsize{$\pm$ 0.2} &
  - &
  12.6 \scriptsize{$\pm$ 0.3} &
  5.0 \scriptsize{$\pm$ 0.3} &
  0.7 \scriptsize{$\pm$ 0.2} \\ \hline
\multirow{2}{*}{\#PraCegoVer 10 Words} &
  CIDEr-D &
  8.6 \scriptsize{$\pm$ 0.1} &
  8.8 \scriptsize{$\pm$ 0.1} &
  - &
  15.0 \scriptsize{$\pm$ 0.9} &
  6.4 \scriptsize{$\pm$ 0.5} &
  1.0 \scriptsize{$\pm$ 0.1} \\
 &
  CIDEr-R &
  9.6 \scriptsize{$\pm$ 0.6} &
  9.3 \scriptsize{$\pm$ 1.1} &
  - &
  15.1 \scriptsize{$\pm$ 0.5} &
  6.2 \scriptsize{$\pm$ 0.3} &
  1.1 \scriptsize{$\pm$ 0.1} \\ \hline
\multirow{2}{*}{\#PraCegoVer 20 Words} &
  CIDEr-D &
  10.2 \scriptsize{$\pm$ 0.5} &
  9.5 \scriptsize{$\pm$ 0.4} &
  - &
  15.9 \scriptsize{$\pm$ 0.1} &
  7.9 \scriptsize{$\pm$ 0.1} &
  1.8 \scriptsize{$\pm$ 0.2} \\
 &
  CIDEr-R &
  16.1 \scriptsize{$\pm$ 0.8} &
  10.9 \scriptsize{$\pm$ 0.6} &
  - &
  15.4 \scriptsize{$\pm$ 0.3} &
  6.7 \scriptsize{$\pm$ 0.0} &
  1.7 \scriptsize{$\pm$ 0.1} \\ \hline

\end{tabular}%
}
\end{table*}

Overall, the models' performance decreases as the variance of the reference length rises. Considering the datasets MS COCO, MS COCO -- Subset (Multi-Reference), and Flickr30K, in which multiple references are available, one can compare the models' performance using the SPICE score, since it has a high human correlation, to avoid a casual bias. We can observe that as the reference length variance increases, the performance of the models optimizing CIDEr-D drops faster than those that optimize CIDEr-R. In particular, the performances, based on the SPICE score, of the models optimizing CIDEr-D and CIDEr-R are not statistically different on Flickr30K, whose reference lengths have the highest standard deviation. Moreover, the models' performance on MS COCO -- Subset (Single-Reference) is worse than on MS COCO -- Subset (Multi-Reference), which shows the negative impact of lack of multiple references for each example. Still, it can be noted that in the case of MS COCO -- Subset (Single-Reference), the model that optimized CIDEr-R surpassed the one that optimized CIDEr-D. It means that the optimization of CIDEr-R is preferred in the case of a Single Reference.

Concerning the subsets of \#PraCegoVer, the metrics ROUGE-L, METEOR, and BLEU-4 have low correlation with human judgments, and thus the comparisons must be made in terms of CIDEr-D and \text{CIDEr-D}. That said, one can see that, regarding \#PraCegoVer's subsets, the performance also improves as the variance of the sentence length decreases. Still, in \textit{\#PraCegoVer 20 Words}, we doubled the average sentence length and also rose the variance of length compared to \textit{\#PraCegoVer 10 Words}. For this subset, optimizing CIDEr-R resulted in better models than using CIDEr-D. In the case of \textit{\#PraCegoVer 10 Words} and \textit{\#PraCegoVer 20 Words} datasets, even the average \text{CIDEr-D} score improves when we optimize CIDEr-R. It demonstrates that CIDEr-R is more suitable for this context. Moreover, optimizing CIDEr-R improved the captions' quality in comparison with \text{CIDEr-D} for all subsets of \#PraCegoVer as detailed in Section~\ref{sub:exploratory_analysis}.

\subsection{Qualitative Analysis}
\label{sub:exploratory_analysis}

To qualitatively explore the metrics' weakness, we created the triplets shown in Table \ref{tab:test_metrics}. The triplets are composed of a reference sentence and two candidates: \textit{Correct Candidate} correctly describes the image and \textit{Incorrect Candidate} does not describe the objects in the image. We compute the metrics' scores for each candidate, and then we compute the accuracy based on these triplets, as explained in Section~\ref{sub:human_level_correlation}. We constructed the first example to evaluate the performance of the metrics when the Incorrect Candidate corresponds to a change of the subject of the reference caption, thus changing its meaning entirely, whereas the Correct Candidate consists of small syntactic changes but keeping the same meaning. In the second example, we explore different sentences for the Correct Candidate, while the Incorrect Candidate has some n-grams matching with the reference but has many word repetitions and is semantically incomplete. In the third example, we created the triplets to explore differences in length between the candidates and the reference. Thus, we selected a small sentence for the Correct Candidate, and we constructed the Incorrect Candidates by removing parts of the reference caption and replacing the dog with a cat. Still, in one of the Incorrect Candidates, we repeated the word distance to increase the length.\vspace{-0.01cm}

SPICE beats all other metrics, assigning a higher score for the Correct Candidate in 11 out of 15 triplets. However, it is worth noting that it misclassified triplets in which words are repeated. SPICE can reach this performance because it parses the descriptions according to linguistic rules, limiting its extension to other languages because it depends on linguistic experts. Regarding the other metrics, one can note that BLEU-4 does not classify correctly any triplets, whereas CIDEr-D and ROUGE-L assigned the higher score to the Correct Candidate when its sentence syntactically equals the Incorrect Candidate except by the word ``cat''. Similarly, CIDEr-R and METEOR correctly classified 4 triplets. METEOR has better performance when the candidate and reference sentences are more syntactically similar. In contrast, \text{CIDEr-R} correctly penalizes sentences with word repetition and still while assigning the Correct Candidate. Also, it is essential to highlight that CIDEr-R is more invariant to sentence length than CIDEr-D. We saw this in cases where a higher score is assigned to the Correct Candidate even though it is shorter than the Incorrect Candidate. For example, in the pair ``A cow and a dog on the street.'' and ``A bird sitting on the on on on cat and a dog and a dog and and sitting on next'', the first candidate is preferred.

\begin{table*}[!t]
\caption{A small set of triplets created to explore the  metrics' weakness. We constructed 5 triplets for each image, each one consisting of a reference, \textit{Correct Candidate (CC)} and \textit{Incorrect Candidate (IC)}. \textit{Correct Candidate} correctly describes the image, while \textit{Incorrect Candidate} is a sentence that do not describes the image.}\vspace{-0.1cm}
\label{tab:test_metrics}
\resizebox{\textwidth}{!}{%
\begin{tabular}{ll|l|cc|cc|cc|cc|cc|cc|}
\hline
\multicolumn{1}{|c|}{\multirow{2}{*}{\textbf{Reference}}} &
  \multicolumn{1}{c|}{\multirow{2}{*}{\textbf{Correct Candidate (CC)}}} &
  \multicolumn{1}{c|}{\multirow{2}{*}{\textbf{Incorrect Candidate (IC)}}} &
  \multicolumn{2}{c|}{\textbf{CIDEr-R}} &
  \multicolumn{2}{c|}{\textbf{CIDEr-D}} &
  \multicolumn{2}{c|}{\textbf{SPICE}} &
  \multicolumn{2}{c|}{\textbf{METEOR}} &
  \multicolumn{2}{c|}{\textbf{ROUGE-L}} &
  \multicolumn{2}{c|}{\textbf{BLEU-4}} \\
\multicolumn{1}{|c|}{} &
  \multicolumn{1}{c|}{} &
  \multicolumn{1}{c|}{} &
  \textbf{CC} &
  \textbf{IC} &
  \textbf{CC} &
  \textbf{IC} &
  \textbf{CC} &
  \textbf{IC} &
  \textbf{CC} &
  \textbf{IC} &
  \textbf{CC} &
  \textbf{IC} &
  \textbf{CC} &
  \textbf{IC} \\ \hline
\multicolumn{1}{|l|}{\multirow{5}{*}{\begin{tabular}[c]{@{}l@{}} \\ \hspace{2.5cm} \includegraphics[height=0.8in,width=1.0in]{mscoco_girl.jpg} \\ 1. A young girl is preparing to blow out her \\ candle. \end{tabular}}} &
  A young girl is about to blow out her candle. &
  \multirow{5}{*}{\begin{tabular}[c]{@{}l@{}}A young dog is preparing to blow \\ out her candle.\end{tabular}} &
  452.3 &
  \textbf{537.4} &
  452.3 &
  \textbf{537.4} &
  \textbf{80.0} &
  20.0 &
  \textbf{92.5} &
  47.2 &
  90.0 &
  \textbf{90.0} &
  65.8 &
  \textbf{70.7} \\
\multicolumn{1}{|l|}{} &
  \begin{tabular}[c]{@{}l@{}}A young girl is getting ready to blow out \\ her candle.\end{tabular} &
   &
  389.5 &
  \textbf{537.4} &
  384.9 &
  \textbf{537.4} &
  \textbf{80.0} &
  20.0 &
  \textbf{53.6} &
  47.2 &
  86.5 &
  \textbf{90.0} &
  58.8 &
  \textbf{70.7} \\
\multicolumn{1}{|l|}{} &
  \begin{tabular}[c]{@{}l@{}}A young girl is getting ready to blow \\ out a candle.\end{tabular} &
   &
  238.1 &
  \textbf{537.4} &
  248.7 &
  \textbf{537.4} &
  \textbf{66.7} &
  20.0 &
  46.6 &
  \textbf{47.2} &
  76.9 &
  \textbf{90.0} &
  35.1 &
  \textbf{70.7} \\
\multicolumn{1}{|l|}{} &
  \begin{tabular}[c]{@{}l@{}}A young girl is getting ready to blow \\ out a candle on a small dessert.\end{tabular} &
   &
  173.6 &
  \textbf{537.4} &
  138.0 &
  \textbf{537.4} &
  \textbf{50.0} &
  20.0 &
  44.1 &
  \textbf{47.2} &
  66.4 &
  \textbf{90.0} &
  24.6 &
  \textbf{70.7} \\
\multicolumn{1}{|l|}{} &
  \begin{tabular}[c]{@{}l@{}}A kid is to blow out the single candle in a \\ bowl of birthday goodness.\end{tabular} &
   &
  78.7 &
  \textbf{537.4} &
  60.8 &
  \textbf{537.4} &
  \textbf{15.4} &
  20.0 &
  25.6 &
  \textbf{47.2} &
  49.8 &
  \textbf{90.0} &
  0.0 &
  \textbf{70.7} \\ \hline
\multicolumn{1}{|l|}{\multirow{5}{*}{\begin{tabular}[c]{@{}l@{}}\hspace{2.5cm}  \includegraphics[height=0.8in,width=1.0in]{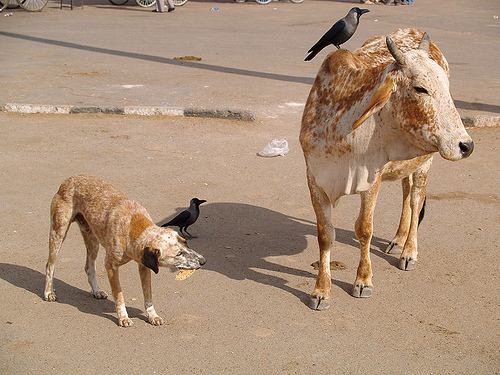} \\2. A bird sitting on the back of a cow and a dog \\ and bird standing on the ground next to the cow.\\  \end{tabular}}} &
  \begin{tabular}[c]{@{}l@{}}A mottled brown dog and cow with two \\ little birds outdoors.\end{tabular} &
  \multirow{5}{*}{\begin{tabular}[c]{@{}l@{}}A bird sitting on the on on on cat and a \\ dog and a dog and and sitting on next to.\end{tabular}} &
  32.1 &
  \textbf{129.6} &
  7.5 &
  \textbf{201.1} &
  10.5 &
  \textbf{15.4} &
  8.8 &
  \textbf{25.9} &
  22.9 &
  \textbf{60.2} &
  0.0 &
  \textbf{30.8} \\
\multicolumn{1}{|l|}{} &
 A cow standing next to a dog on dirt ground. &
   &
  89.1 &
  \textbf{129.6} &
  15.4 &
  \textbf{201.1} &
  \textbf{35.3} &
  15.4 &
  21.8 &
  \textbf{25.9} &
  35.1 &
  \textbf{60.2} &
  0.0 &
  \textbf{30.8} \\
\multicolumn{1}{|l|}{} &
  A dog with a bird and a large cow on a street. &
   &
  73.2 &
  \textbf{129.6} &
  25.9 &
  \textbf{201.1} &
  \textbf{22.2} &
  15.4 &
  15.6 &
  \textbf{25.9} &
  27.9 &
  \textbf{60.2} &
  0.0 &
  \textbf{30.8} \\
\multicolumn{1}{|l|}{} &
  \begin{tabular}[c]{@{}l@{}}A dog and a cow with a bird sitting on \\ its back.\end{tabular} &
   &
  \textbf{168.7} &
  129.6 &
  54.7 &
  \textbf{201.1} &
  \textbf{35.3} &
  15.4 &
  \textbf{26.7} &
  25.9 &
  33.5 &
  \textbf{60.2} &
  15.8 &
  \textbf{30.8} \\
\multicolumn{1}{|l|}{} &
  A cow and a dog on a street. &
   &
  \textbf{133.2} &
  129.6 &
  12.5 &
  \textbf{201.1} &
  14.3 &
  \textbf{15.4} &
  14.4 &
  \textbf{25.9} &
  36.9 &
  \textbf{60.2} &
  9.8 &
  \textbf{30.8} \\ \hline
\multicolumn{1}{|l|}{\multirow{5}{*}{\begin{tabular}[c]{@{}l@{}}\hspace{2.5cm} \includegraphics[height=0.8in,width=1.0in]{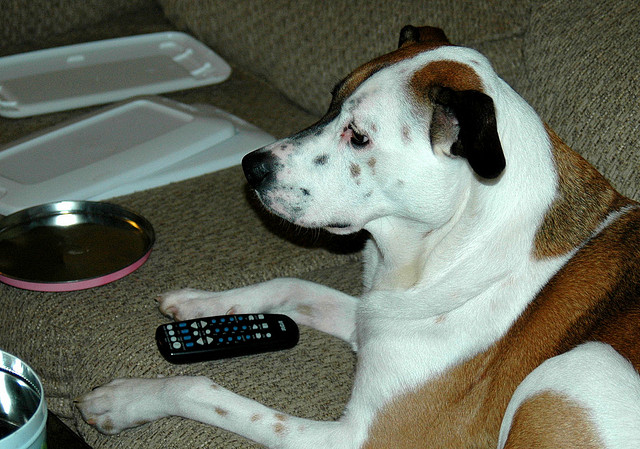} \\ 3. Dog laying on couch looking into distance \\ with remote control by paw \\ \end{tabular}}} &
  \multirow{5}{*}{A dog laying with a remote control.} &
  \begin{tabular}[c]{@{}l@{}}Cat laying on couch looking into distance \\ with remote control by paw.\end{tabular} &
  160.1 &
  \textbf{745.3} &
  128.5 &
  \textbf{745.3} &
  50.0 &
  \textbf{88.9} &
  22.9 &
  \textbf{56.1} &
  50.2 &
  \textbf{91.7} &
  0.0 &
  \textbf{90.4} \\
\multicolumn{1}{|l|}{} &
   &
  Cat looking into distance with remote control. &
  160.1 &
  \textbf{415.8} &
  128.5 &
  \textbf{304.2} &
  \textbf{50.0} &
  46.2 &
  22.9 &
  \textbf{30.9} &
  50.2 &
  \textbf{60.3} &
  0.0 &
  \textbf{39.6} \\
\multicolumn{1}{|l|}{} &
   &
  \begin{tabular}[c]{@{}l@{}}Cat looking into distance distance distance \\ with a remote control.\end{tabular} &
  \textbf{160.1} &
  142.8 &
  128.5 &
  \textbf{151.6} &
  50.0 &
  \textbf{53.3} &
  22.9 &
  \textbf{24.4} &
  50.2 &
  \textbf{53.7} &
  0.0 &
  0.0 \\
\multicolumn{1}{|l|}{} &
   &
  Cat with remote control by paw. &
  160.1 &
  \textbf{332.5} &
  128.5 &
  \textbf{212.0} &
  \textbf{50.0} &
  40.0 &
  22.9 &
  \textbf{24.0} &
  50.2 &
  \textbf{52.4} &
  0.0 &
  \textbf{28.0} \\
\multicolumn{1}{|l|}{} &
   &
  A cat with a remote control. &
  \textbf{160.1} &
  77.0 &
  \textbf{128.5} &
  54.8 &
  \textbf{50.0} &
  30.8 &
  \textbf{22.9} &
  12.2 &
  \textbf{50.2} &
  31.4 &
  0.0 &
  0.0 \\ \hline
\multicolumn{1}{c}{} &
  \multicolumn{1}{c|}{} &
  \multicolumn{1}{r|}{\textbf{Accuracy:}} &
  \multicolumn{2}{c|}{4/15} &
  \multicolumn{2}{c|}{1/15} &
  \multicolumn{2}{c|}{11/15} &
  \multicolumn{2}{c|}{4/15} &
  \multicolumn{2}{c|}{1/15} &
  \multicolumn{2}{c|}{0/15} \\ \cline{3-15} 
\end{tabular}%
}
\end{table*}
\begin{figure*}[!t]
\centering
    \subfloat[
    \textbf{Reference}:  ``Na foto, Thalita Gelenske e Thaís Silva estão abraçadas com Luana Génot na livraria Travessa. Ao fundo, diversos livros coloridos estão na prateleira. Nas laterais da foto, existem 2 banners: um deles vermelho, com o logo da, e o outro com a divulgação do livro da Luana.''\\
    \textbf{CIDEr-D}: ``Foto de uma mulher segurando um livro com \textcolor{blue}{\textit{livros}}. Ao fundo, uma estante com \textcolor{blue}{\textit{livros}}. Texto: “A \textcolor{blue}{\textit{sua}}. \textcolor{blue}{\textit{É sua}} festa. \textcolor{blue}{\textit{É sua}}!”.''\\
    \textbf{CIDEr-R}: ``Fotografia colorida. Bárbara está sentada em uma biblioteca. Ela está com uma menina. Ela veste uma blusa branca. Ao fundo, uma estante de livros.''.]{\includegraphics[width=0.32\textwidth,trim={0 0 0 0cm} ,clip]{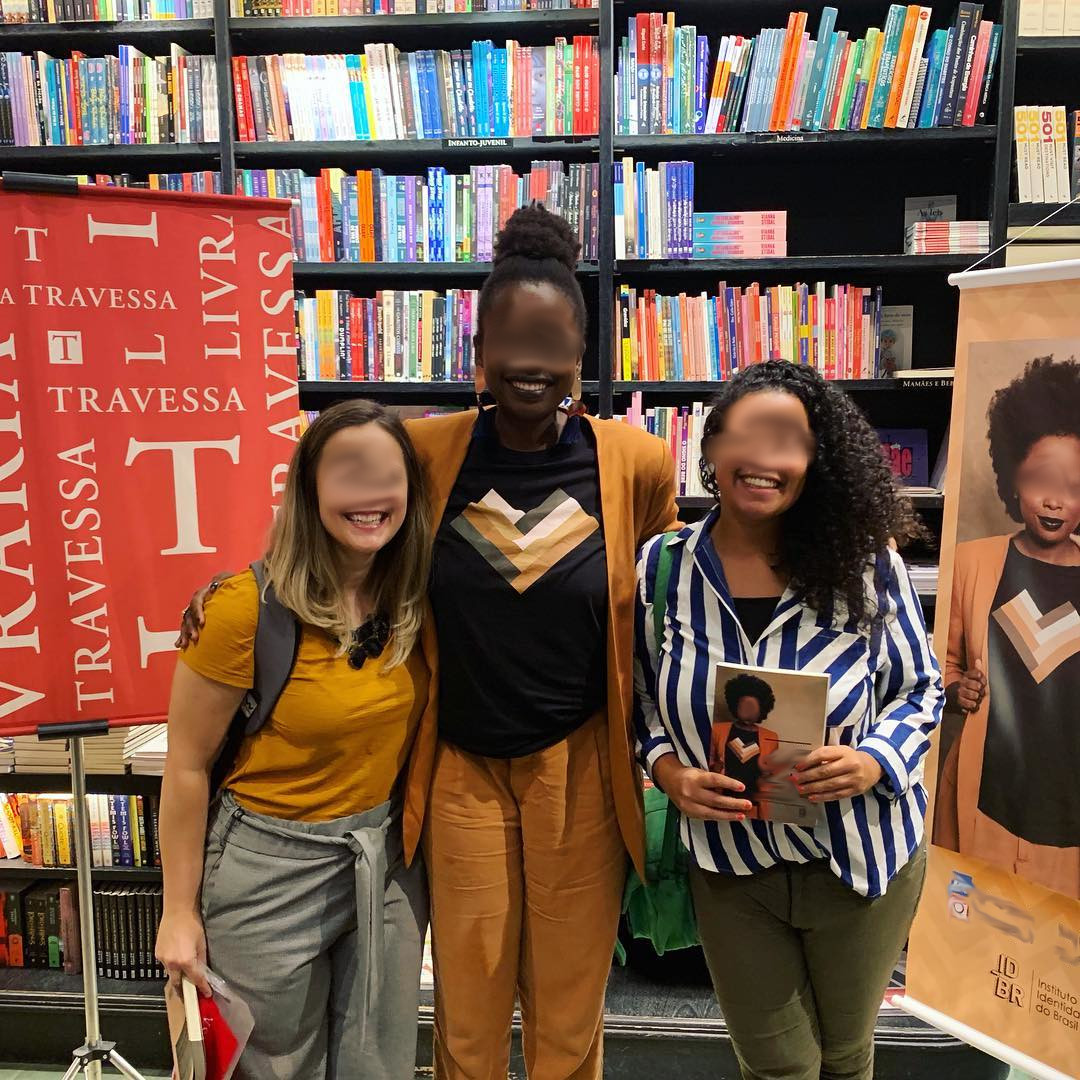}\label{fig:generated_descriptions_library}}\hspace{0.2cm}
    \subfloat[
    \textbf{Reference}: ``Two kayaks are sitting on a river bank empty.''\\
    \textbf{Subset Single Ref. CIDEr-D}: ``a \textcolor{blue}{\textit{yellow}} and \textcolor{blue}{\textit{yellow}} \textcolor{blue}{\textit{boat}} sitting on the side of a beach next to a \textcolor{blue}{\textit{boat}}'\\
    \textbf{Subset Single Ref. CIDEr-R}: ``a yellow surfboard sitting on the beach next to the ocean''\\
    ]{\includegraphics[width=0.32\textwidth]{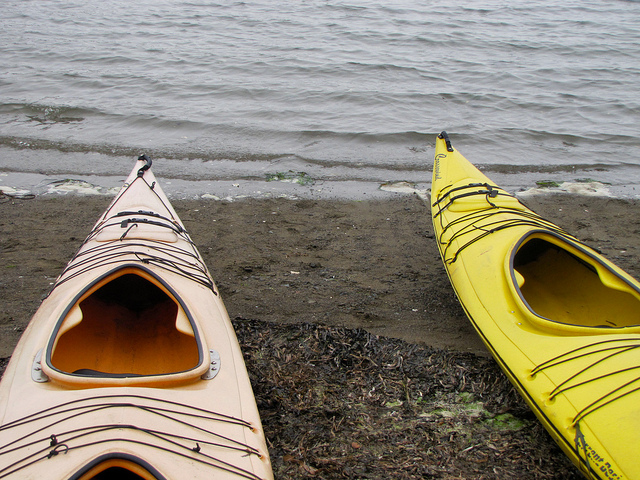}\label{fig:generated_caption_kayak}}\hspace{0.2cm}
    \subfloat[
    \textbf{Reference}: ``A cat stares at itself in a mirror.''\\
    \textbf{Subset Single Ref. CIDEr-D}: ``a black and white cat sitting in a \textcolor{blue}{\textit{mirror}} looking in a \textcolor{blue}{\textit{mirror}}'\\
    \textbf{Subset Single Ref. CIDEr-R}: ``a cat sitting in front of a mirror''\\
    ]{\includegraphics[width=0.32\textwidth]{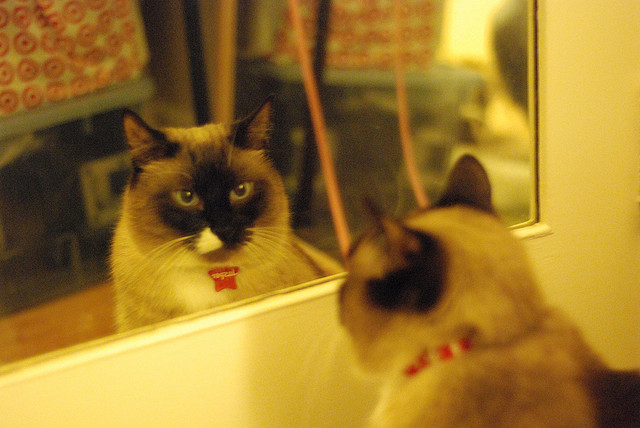}\label{fig:generated_caption_cat}}
    \caption{Figure \ref{fig:generated_descriptions_library} is an example of image from \textit{\#PraCegoVer-63K} test set and Figures \ref{fig:generated_caption_kayak} and \ref{fig:generated_caption_cat} are from  MS COCO Karpathy test set \cite{karpathy2015deep}. In Figure \ref{fig:generated_descriptions_library}, we show the reference caption and the description generated by models trained on \textit{\#PraCegoVer-63K} to optimize CIDEr-D and CIDEr-R. In Figures \ref{fig:generated_caption_kayak} and \ref{fig:generated_caption_cat}, we show the captions generated by  models trained on MS COCO -- Subset (Single-Reference) to optimize CIDEr-D and CIDEr-R, as well as the available reference for that image. ``Subset Single Ref.'' stands for MS COCO -- Subset (Single-Reference).}
    \label{fig:generated_descriptions}
\end{figure*} 

Still, to show the quality of the captions generated by models trained to optimize CIDEr-D and CIDEr-R in the context of a single reference, we present in Figure \ref{fig:generated_descriptions} two examples from \textit{\#PraCegoVer-63K} test set and two from MS COCO. In these examples, the images are followed by the reference caption and the descriptions automatically generated by the models. Regarding MS COCO, we train the models on MS COCO -- Subset (Single-Reference), and we generate the captions for images from MS COCO Karpathy test set \cite{karpathy2015deep}.

Models trained to optimize CIDEr-D learn to repeat words in all cases, as highlighted in Figure \ref{fig:generated_descriptions}. This occurs because the length penalty in CIDEr-R is more flexible than the Gaussian penalty in CIDEr-D. Hence, the model can generate short captions without being much penalized. As a result, the models trained to optimize CIDEr-R tend to produce summarized descriptions. This become clear specially when we consider \textit{\#PraCegoVer-63K}  where the references are long, as illustrated in Figures \ref{fig:generated_descriptions_city1} and \ref{fig:generated_descriptions_library}. Therefore, CIDEr-R is a better metric to be optimized in the context of long reference captions, especially when the variance of the sentences is high.

%% file: 07-conclusion.tex
\section{Conclusion}
\label{sec:conclusion}
This work demonstrated that CIDEr-D's performance is hampered by the lack of multiple reference sentences and sentence lengths variance. Thus, we proposed CIDEr-R as an alternative to CIDEr-D in the context of a single reference and where the sentence length variance matters, such as large datasets based on social media data. We modified the Gaussian penalty to make CIDEr-D invariant to the difference in length between the reference caption and the generated description. Also, we introduced a penalty for word repetition to better compare captions with a length similar to the reference but with words repeated several times. Further, we showed that CIDEr-R is more accurate and closer to human judgment than CIDEr-D, and it is also more robust regarding the number of references. Still, we showed that CIDEr-R is a preferred evaluation metric to CIDEr-D for datasets with single reference and high variance of sentence length, such as \#PraCegoVer. Additionally, we demonstrated that using SCST to optimize CIDEr-D is not a good approach regarding datasets with a high length variance because the models are more likely to repeat words to increase the length of the generated caption. Finally, we demonstrated that training models to optimize CIDEr-R instead of \text{CIDEr-D} produces more descriptive captions.

For future work, we intend to create a dataset similar to PASCAL-50S with captions in Portugue\-se with the same characteristics of \#PraCegoVer to validate the accuracy of the metrics in the context of long sentences and high variance. Also, we intend to remove proper names and process synonyms to reduce the vocabulary size and replace words with low-frequency with synonyms more common in subsets of \#PraCegoVer to evaluate how this impacts the models' performance.

%% file: emnlp2021.bbl
\begin{thebibliography}{34}
\expandafter\ifx\csname natexlab\endcsname\relax\def\natexlab#1{#1}\fi

\bibitem[{Agrawal et~al.(2019)Agrawal, Anderson, Desai, Wang, Chen, Jain,
  Johnson, Batra, Parikh, and Lee}]{agrawal2019nocaps}
Harsh Agrawal, Peter Anderson, Karan Desai, Yufei Wang, Xinlei Chen, Rishabh
  Jain, Mark Johnson, Dhruv Batra, Devi Parikh, and Stefan Lee. 2019.
\newblock \href {https://doi.org/10.1109/ICCV.2019.00904} {nocaps: novel object
  captioning at scale}.
\newblock In \emph{2019 {IEEE/CVF} International Conference on Computer Vision,
  {ICCV} 2019, Seoul, Korea (South), October 27 - November 2, 2019}, pages
  8947--8956. {IEEE}.

\bibitem[{Anderson et~al.(2016)Anderson, Fernando, Johnson, and
  Gould}]{anderson2016spice}
Peter Anderson, Basura Fernando, Mark Johnson, and Stephen Gould. 2016.
\newblock Spice: Semantic propositional image caption evaluation.
\newblock In \emph{European Conference on Computer Vision}, pages 382--398.

\bibitem[{Anderson et~al.(2018)Anderson, He, Buehler, Teney, Johnson, Gould,
  and Zhang}]{anderson2018bottom}
Peter Anderson, Xiaodong He, Chris Buehler, Damien Teney, Mark Johnson, Stephen
  Gould, and Lei Zhang. 2018.
\newblock \href {https://doi.org/10.1109/CVPR.2018.00636} {Bottom-up and
  top-down attention for image captioning and visual question answering}.
\newblock In \emph{2018 {IEEE} Conference on Computer Vision and Pattern
  Recognition, {CVPR} 2018, Salt Lake City, UT, USA, June 18-22, 2018}, pages
  6077--6086. {IEEE} Computer Society.

\bibitem[{Banerjee and Lavie(2005)}]{METEOR2005}
Satanjeev Banerjee and Alon Lavie. 2005.
\newblock \href {https://aclanthology.org/W05-0909} {{METEOR}: An automatic
  metric for {MT} evaluation with improved correlation with human judgments}.
\newblock In \emph{Proceedings of the {ACL} Workshop on Intrinsic and Extrinsic
  Evaluation Measures for Machine Translation and/or Summarization}, pages
  65--72, Ann Arbor, Michigan. Association for Computational Linguistics.

\bibitem[{Bergstra and Bengio(2012)}]{bergstra2012random}
James Bergstra and Yoshua Bengio. 2012.
\newblock Random search for hyper-parameter optimization.
\newblock \emph{The Journal of Machine Learning Research}, 13(1):281--305.

\bibitem[{Chen et~al.(2015)Chen, Fang, Lin, Vedantam, Gupta, Doll{\'a}r, and
  Zitnick}]{chen2015microsoft}
Xinlei Chen, Hao Fang, Tsung-Yi Lin, Ramakrishna Vedantam, Saurabh Gupta, Piotr
  Doll{\'a}r, and C~Lawrence Zitnick. 2015.
\newblock \href {https://arxiv.org/abs/1504.00325} {Microsoft coco captions:
  Data collection and evaluation server}.
\newblock \emph{arXiv preprint arXiv:1504.00325}.

\bibitem[{Cornia et~al.(2020)Cornia, Stefanini, Baraldi, and
  Cucchiara}]{cornia2020meshed}
Marcella Cornia, Matteo Stefanini, Lorenzo Baraldi, and Rita Cucchiara. 2020.
\newblock \href {https://doi.org/10.1109/CVPR42600.2020.01059} {Meshed-memory
  transformer for image captioning}.
\newblock In \emph{2020 {IEEE/CVF} Conference on Computer Vision and Pattern
  Recognition, {CVPR} 2020, Seattle, WA, USA, June 13-19, 2020}, pages
  10575--10584. {IEEE}.

\bibitem[{Gurari et~al.(2020)Gurari, Zhao, Zhang, and
  Bhattacharya}]{gurari2020captioning}
Danna Gurari, Yinan Zhao, Meng Zhang, and Nilavra Bhattacharya. 2020.
\newblock Captioning images taken by people who are blind.
\newblock In \emph{European Conference on Computer Vision}, pages 417--434.

\bibitem[{Hodosh et~al.(2013)Hodosh, Young, and Hockenmaier}]{Flickr8K2013}
Micah Hodosh, Peter Young, and Julia Hockenmaier. 2013.
\newblock \href {http://dl.acm.org/citation.cfm?id=2566972.2566993} {Framing
  image description as a ranking task: Data, models and evaluation metrics}.
\newblock \emph{Journal of Artificial Intelligence Research}, 47(1):853--899.

\bibitem[{Huang et~al.(2019)Huang, Wang, Chen, and Wei}]{AoANet_2019}
Lun Huang, Wenmin Wang, Jie Chen, and Xiaoyong Wei. 2019.
\newblock \href {https://doi.org/10.1109/ICCV.2019.00473} {Attention on
  attention for image captioning}.
\newblock In \emph{2019 {IEEE/CVF} International Conference on Computer Vision,
  {ICCV} 2019, Seoul, Korea (South), October 27 - November 2, 2019}, pages
  4633--4642. {IEEE}.

\bibitem[{Johnson et~al.(2016)Johnson, Karpathy, and
  Fei{-}Fei}]{johnson2016densecap}
Justin Johnson, Andrej Karpathy, and Li~Fei{-}Fei. 2016.
\newblock \href {https://doi.org/10.1109/CVPR.2016.494} {Densecap: Fully
  convolutional localization networks for dense captioning}.
\newblock In \emph{2016 {IEEE} Conference on Computer Vision and Pattern
  Recognition, {CVPR} 2016, Las Vegas, NV, USA, June 27-30, 2016}, pages
  4565--4574. {IEEE} Computer Society.

\bibitem[{Karpathy and Li(2015)}]{karpathy2015deep}
Andrej Karpathy and Fei{-}Fei Li. 2015.
\newblock \href {https://doi.org/10.1109/CVPR.2015.7298932} {Deep
  visual-semantic alignments for generating image descriptions}.
\newblock In \emph{{IEEE} Conference on Computer Vision and Pattern
  Recognition, {CVPR} 2015, Boston, MA, USA, June 7-12, 2015}, pages
  3128--3137. {IEEE} Computer Society.

\bibitem[{Kulkarni et~al.(2011)Kulkarni, Premraj, Dhar, Li, Choi, Berg, and
  Berg}]{Kulkarni11babytalk}
Girish Kulkarni, Visruth Premraj, Sagnik Dhar, Siming Li, Yejin Choi,
  Alexander~C Berg, and Tamara~L Berg. 2011.
\newblock Baby talk: Understanding and generating image descriptions.
\newblock In \emph{IEEE Conference on Computer Vision and Pattern Recognition}.

\bibitem[{LeCun et~al.(2015)LeCun, Bengio, and Hinton}]{lecun2015deep}
Yann LeCun, Yoshua Bengio, and Geoffrey Hinton. 2015.
\newblock \href {https://doi.org/10.1038/nature14539} {Deep learning}.
\newblock \emph{nature}, 521(7553):436--444.

\bibitem[{Li et~al.(2020)Li, Yin, Li, Zhang, Hu, Zhang, Wang, Hu, Dong, Wei
  et~al.}]{li2020oscar}
Xiujun Li, Xi~Yin, Chunyuan Li, Pengchuan Zhang, Xiaowei Hu, Lei Zhang, Lijuan
  Wang, Houdong Hu, Li~Dong, Furu Wei, et~al. 2020.
\newblock Oscar: Object-semantics aligned pre-training for vision-language
  tasks.
\newblock In \emph{European Conference on Computer Vision}, pages 121--137.

\bibitem[{Lin(2004)}]{ROUGE2004}
Chin-Yew Lin. 2004.
\newblock \href {https://aclanthology.org/W04-1013} {{ROUGE}: A package for
  automatic evaluation of summaries}.
\newblock In \emph{Text Summarization Branches Out}, pages 74--81, Barcelona,
  Spain. Association for Computational Linguistics.

\bibitem[{Lin et~al.(2014)Lin, Maire, Belongie, Hays, Perona, Ramanan,
  Doll{\'a}r, and Zitnick}]{MSCoco2014}
Tsung-Yi Lin, Michael Maire, Serge Belongie, James Hays, Pietro Perona, Deva
  Ramanan, Piotr Doll{\'a}r, and C~Lawrence Zitnick. 2014.
\newblock Microsoft coco: Common objects in context.
\newblock In \emph{European Conference on Computer Vision}, pages 740--755.

\bibitem[{Liu et~al.(2017)Liu, Zhu, Ye, Guadarrama, and
  Murphy}]{liu2017improved}
Siqi Liu, Zhenhai Zhu, Ning Ye, Sergio Guadarrama, and Kevin Murphy. 2017.
\newblock \href {https://doi.org/10.1109/ICCV.2017.100} {Improved image
  captioning via policy gradient optimization of spider}.
\newblock In \emph{{IEEE} International Conference on Computer Vision, {ICCV}
  2017, Venice, Italy, October 22-29, 2017}, pages 873--881. {IEEE} Computer
  Society.

\bibitem[{Lu et~al.(2018)Lu, Yang, Batra, and Parikh}]{Lu_2018_CVPR}
Jiasen Lu, Jianwei Yang, Dhruv Batra, and Devi Parikh. 2018.
\newblock \href {https://doi.org/10.1109/CVPR.2018.00754} {Neural baby talk}.
\newblock In \emph{2018 {IEEE} Conference on Computer Vision and Pattern
  Recognition, {CVPR} 2018, Salt Lake City, UT, USA, June 18-22, 2018}, pages
  7219--7228. {IEEE} Computer Society.

\bibitem[{Pan et~al.(2020)Pan, Yao, Li, and Mei}]{pan2020x}
Yingwei Pan, Ting Yao, Yehao Li, and Tao Mei. 2020.
\newblock \href {https://doi.org/10.1109/CVPR42600.2020.01098} {X-linear
  attention networks for image captioning}.
\newblock In \emph{2020 {IEEE/CVF} Conference on Computer Vision and Pattern
  Recognition, {CVPR} 2020, Seattle, WA, USA, June 13-19, 2020}, pages
  10968--10977. {IEEE}.

\bibitem[{Papineni et~al.(2002)Papineni, Roukos, Ward, and Zhu}]{BLEU2002}
Kishore Papineni, Salim Roukos, Todd Ward, and Wei-Jing Zhu. 2002.
\newblock \href {https://doi.org/10.3115/1073083.1073135} {{B}leu: a method for
  automatic evaluation of machine translation}.
\newblock In \emph{Proceedings of the 40th Annual Meeting of the Association
  for Computational Linguistics}, pages 311--318, Philadelphia, Pennsylvania,
  USA. Association for Computational Linguistics.

\bibitem[{Pedersoli et~al.(2017)Pedersoli, Lucas, Schmid, and
  Verbeek}]{pedersoli2017areas}
Marco Pedersoli, Thomas Lucas, Cordelia Schmid, and Jakob Verbeek. 2017.
\newblock \href {https://doi.org/10.1109/ICCV.2017.140} {Areas of attention for
  image captioning}.
\newblock In \emph{{IEEE} International Conference on Computer Vision, {ICCV}
  2017, Venice, Italy, October 22-29, 2017}, pages 1251--1259. {IEEE} Computer
  Society.

\bibitem[{Plummer et~al.(2015)Plummer, Wang, Cervantes, Caicedo, Hockenmaier,
  and Lazebnik}]{FlicKr30K2017}
Bryan~A. Plummer, Liwei Wang, Chris~M. Cervantes, Juan~C. Caicedo, Julia
  Hockenmaier, and Svetlana Lazebnik. 2015.
\newblock \href {https://doi.org/10.1109/ICCV.2015.303} {Flickr30k entities:
  Collecting region-to-phrase correspondences for richer image-to-sentence
  models}.
\newblock In \emph{2015 {IEEE} International Conference on Computer Vision,
  {ICCV} 2015, Santiago, Chile, December 7-13, 2015}, pages 2641--2649. {IEEE}
  Computer Society.

\bibitem[{Rennie et~al.(2017)Rennie, Marcheret, Mroueh, Ross, and
  Goel}]{rennie2017self}
Steven~J. Rennie, Etienne Marcheret, Youssef Mroueh, Jerret Ross, and Vaibhava
  Goel. 2017.
\newblock \href {https://doi.org/10.1109/CVPR.2017.131} {Self-critical sequence
  training for image captioning}.
\newblock In \emph{2017 {IEEE} Conference on Computer Vision and Pattern
  Recognition, {CVPR} 2017, Honolulu, HI, USA, July 21-26, 2017}, pages
  1179--1195. {IEEE} Computer Society.

\bibitem[{Santos et~al.(2021)Santos, Colombini, and
  Avila}]{santos2021pracegover}
Gabriel Oliveira~dos Santos, Esther~Luna Colombini, and Sandra Avila. 2021.
\newblock \href {https://arxiv.org/abs/2103.11474} {\#{PraCegoVer: A} large
  dataset for image captioning in portuguese}.
\newblock \emph{arXiv preprint arXiv:2103.11474}.

\bibitem[{Schuster et~al.(2015)Schuster, Krishna, Chang, Fei-Fei, and
  Manning}]{schuster2015generating}
Sebastian Schuster, Ranjay Krishna, Angel Chang, Li~Fei-Fei, and Christopher~D.
  Manning. 2015.
\newblock \href {https://doi.org/10.18653/v1/W15-2812} {Generating semantically
  precise scene graphs from textual descriptions for improved image retrieval}.
\newblock In \emph{Proceedings of the Fourth Workshop on Vision and Language},
  pages 70--80, Lisbon, Portugal. Association for Computational Linguistics.

\bibitem[{Sellam et~al.(2020)Sellam, Das, and Parikh}]{sellam-etal-2020-bleurt}
Thibault Sellam, Dipanjan Das, and Ankur Parikh. 2020.
\newblock \href {https://doi.org/10.18653/v1/2020.acl-main.704} {{BLEURT}:
  Learning robust metrics for text generation}.
\newblock In \emph{Proceedings of the 58th Annual Meeting of the Association
  for Computational Linguistics}, pages 7881--7892, Online. Association for
  Computational Linguistics.

\bibitem[{Vaswani et~al.(2017)Vaswani, Shazeer, Parmar, Uszkoreit, Jones,
  Gomez, Kaiser, and Polosukhin}]{attention_is_all_you_need_2017}
Ashish Vaswani, Noam Shazeer, Niki Parmar, Jakob Uszkoreit, Llion Jones,
  Aidan~N. Gomez, Lukasz Kaiser, and Illia Polosukhin. 2017.
\newblock \href
  {https://proceedings.neurips.cc/paper/2017/hash/3f5ee243547dee91fbd053c1c4a845aa-Abstract.html}
  {Attention is all you need}.
\newblock In \emph{Advances in Neural Information Processing Systems 30: Annual
  Conference on Neural Information Processing Systems 2017, December 4-9, 2017,
  Long Beach, CA, {USA}}, pages 5998--6008.

\bibitem[{Vedantam et~al.(2015)Vedantam, Zitnick, and Parikh}]{CIDEr2015}
Ramakrishna Vedantam, C.~Lawrence Zitnick, and Devi Parikh. 2015.
\newblock \href {https://doi.org/10.1109/CVPR.2015.7299087} {Cider:
  Consensus-based image description evaluation}.
\newblock In \emph{{IEEE} Conference on Computer Vision and Pattern
  Recognition, {CVPR} 2015, Boston, MA, USA, June 7-12, 2015}, pages
  4566--4575. {IEEE} Computer Society.

\bibitem[{Vinyals et~al.(2015)Vinyals, Toshev, Bengio, and
  Erhan}]{vinyals2015show}
Oriol Vinyals, Alexander Toshev, Samy Bengio, and Dumitru Erhan. 2015.
\newblock \href {https://doi.org/10.1109/CVPR.2015.7298935} {Show and tell: {A}
  neural image caption generator}.
\newblock In \emph{{IEEE} Conference on Computer Vision and Pattern
  Recognition, {CVPR} 2015, Boston, MA, USA, June 7-12, 2015}, pages
  3156--3164. {IEEE} Computer Society.

\bibitem[{Wang et~al.(2020)Wang, Huang, Zhang, and Sun}]{wang2020visual}
Tan Wang, Jianqiang Huang, Hanwang Zhang, and Qianru Sun. 2020.
\newblock \href {https://doi.org/10.1109/CVPR42600.2020.01077} {Visual
  commonsense {R-CNN}}.
\newblock In \emph{2020 {IEEE/CVF} Conference on Computer Vision and Pattern
  Recognition, {CVPR} 2020, Seattle, WA, USA, June 13-19, 2020}, pages
  10757--10767. {IEEE}.

\bibitem[{{Web para Todos}(2018)}]{criadoraPracegover}
{Web para Todos}. 2018.
\newblock \href
  {http://mwpt.com.br/criadora-do-projeto-pracegover-incentiva-descricao-de-imagens-na-web}
  {{Criadora do projeto \#PraCegoVer incentiva a descrição de imagens na
  web}}.
\newblock
  http://mwpt.com.br/criadora-do-projeto-pracegover-incentiva-descricao-de-imagens-na-web.

\bibitem[{Zhang et~al.(2020)Zhang, Kishore, Wu, Weinberger, and
  Artzi}]{zhang2019bertscore}
Tianyi Zhang, Varsha Kishore, Felix Wu, Kilian~Q. Weinberger, and Yoav Artzi.
  2020.
\newblock \href {https://openreview.net/forum?id=SkeHuCVFDr} {Bertscore:
  Evaluating text generation with {BERT}}.
\newblock In \emph{8th International Conference on Learning Representations,
  {ICLR} 2020, Addis Ababa, Ethiopia, April 26-30, 2020}. OpenReview.net.

\bibitem[{Zhao et~al.(2019)Zhao, Peyrard, Liu, Gao, Meyer, and
  Eger}]{zhao-etal-2019-moverscore}
Wei Zhao, Maxime Peyrard, Fei Liu, Yang Gao, Christian~M. Meyer, and Steffen
  Eger. 2019.
\newblock \href {https://doi.org/10.18653/v1/D19-1053} {{M}over{S}core: Text
  generation evaluating with contextualized embeddings and earth mover
  distance}.
\newblock In \emph{Proceedings of the 2019 Conference on Empirical Methods in
  Natural Language Processing and the 9th International Joint Conference on
  Natural Language Processing (EMNLP-IJCNLP)}, pages 563--578, Hong Kong,
  China. Association for Computational Linguistics.

\end{thebibliography}
